# Vehicle-in-Virtual-Environment (VVE) Based Autonomous Driving Function Development and Evaluation Methodology for Vulnerable Road User Safety

## Haochong Chen, Xincheng Cao, Levent Guvenc, Bilin Aksun Guvenc

Automated Driving Lab, Ohio State University

## Abstract

Traditional methods for developing and evaluating autonomous driving functions, such as model-in-the-loop (MIL) and hardware-in-the-loop (HIL) simulations, heavily depend on the accuracy of simulated vehicle models and human factors, especially for vulnerable road user safety systems. Continuation of development during public road deployment forces other road users including vulnerable ones to involuntarily participate in the development process, leading to safety risks, inefficiencies, and a decline in public trust. To address these deficiencies, the Vehicle-in-Virtual-Environment (VVE) method was proposed as a safer, more efficient, and cost-effective solution for developing and testing connected and autonomous driving technologies by operating the real vehicle and multiple other actors like vulnerable road users in different test areas while being immersed within the same highly realistic virtual environment. This VVE approach synchronizes real-world vehicle and vulnerable road user motion within the same virtual scenario, enabling the safe and realistic testing of various traffic situations in a safe and repeatable manner. In this paper, we propose a new testing pipeline that sequentially integrates MIL, HIL, and VVE methods to comprehensively develop and evaluate autonomous driving functions. The effectiveness of this testing pipeline will be demonstrated using an autonomous driving path-tracking algorithm with local deep reinforcement learning modification for vulnerable road user collision avoidance.

## Introduction

As urbanization accelerates and technology continues to evolve, the number of privately owned vehicles has steadily increased each year. This surge has led to critical challenges for modern cities, such as traffic congestion and an alarming rise in car accidents. In recent years, research in autonomous driving has been progressing rapidly, with many successful deployments [1] including autonomous shuttles in urban areas [2-3] with the expectation of significantly reducing car accidents caused by human error and the associated injuries and fatalities involving vulnerable road users. According to the World Health Organization's (WHO) Global Status Report on Road Safety, more than 50 million people are injured and 1.3 million annually lost their lives worldwide in car accidents [4]. In the U.S. alone, 2.4 million individuals were injured and nearly 43,000 fatalities occurred due to car accidents [5]. Approximately 75% of these accidents are caused by human errors such as distracted driving, driving under the influence, and drowsy driving [6].

To address these challenges, Automated Driving Systems (ADS) are being developed as a highly promising solution. By taking advantage of robust and high-performance autonomous driving algorithms, ADS have the potential to significantly reduce accidents caused by human error. The Society of Automotive Engineers (SAE) classifies autonomous vehicles into six levels, ranging from fully manual driving (Level 0) to fully autonomous driving (Level 5) [7]. Vehicles at SAE Levels 4 and 5 show great promise in mitigating accidents by relying on advanced and consistent algorithmic performance across various traffic conditions. To increase the level of autonomy, extensive research and testing have been conducted in recent years [5-8]. However, alongside these developments, new challenges and potential issues have also emerged.

Traditional testing pipelines, such as Model-in-the-Loop (MIL) [12] and Hardware-in-the-Loop (HIL) [10-11] simulations followed by deployment on public roads, have significant limitations in terms of safety, cost, and efficiency. These methods often require exposing other road users to autonomous vehicles equipped with unverified or experimental driving functions, raising substantial safety and ethical concerns. Moreover, the dependence on physical road testing is both expensive and time-consuming, which can slow down the iterative development and refinement of autonomous driving technologies.

To address these challenges, a novel approach called Vehicle-in-Virtual-Environment (VVE) was proposed [12-13]. The VVE method integrates real vehicles into highly realistic virtual environments, enabling comprehensive and resource-efficient testing without the need for public road exposure. This approach not only enhances safety by eliminating risks to other road users but also significantly reduces testing costs and accelerates the development process. Additionally, VVE is particularly well-suited for training and evaluating Deep Reinforcement Learning (DRL) based autonomous driving agents. By providing a controlled yet dynamic simulation environment, VVE allows DRL agents to interact with diverse traffic scenarios and adapt to complex, real-world-like conditions safely and efficiently. This integration facilitates the thorough validation of autonomous driving systems, ensuring robust performance and safety before any public road deployment. Consequently, the VVE method offers a safer, more efficient, and cost-effective solution for advancing autonomous driving technology.

In response to the aforementioned deficiencies and challenges, this paper proposes a novel testing pipeline for evaluating autonomous driving decision-making and control algorithms. This comprehensive testing pipeline integrates Model-in-the-Loop (MIL), Hardware-in-the-Loop (HIL), Vehicle-in-Virtual-Environment (VVE) testing, and public road testing to thoroughly validate the algorithms. Initially, the algorithm is developed and tested using the MIL approach. A comprehensive and detailed vehicle dynamics model is created using Simulink, incorporating traditional longitudinal and lateral dynamics, tire rotation, and tire force models. By adjusting parameters, this model can simulate the vehicle's dynamics with high accuracy, allowing for effective preliminary develop and testing. Moreover, deep learning or reinforcement learning based autonomous driving algorithms can also be trained and validated using this MIL testing method. Subsequently, the developed algorithm is evaluated under HIL settings. HIL testing integrates real hardware components with the simulation environment, enabling us to simulate signal transmission delays and eliminate unrealistic extreme inputs, such as abrupt steering changes from left to right. This step brings the algorithm closer to real-world conditions by accounting for hardware-related factors that can affect performance. The VVE testing method is then employed to further evaluate the algorithm. The VVE method synchronizes the real-world motions of an actual vehicle with its virtual counterpart in a simulated environment, allowing for the creation of various virtual traffic scenarios for safe and resource-efficient testing. Because real vehicles and pedestrians participate in the simulation process, the dynamics captured are more authentic, enhancing the credibility of the simulation results. VVE significantly reduces testing costs and time, improving testing efficiency by enabling extensive scenario testing without the risks associated with public road exposure. In addition, the VVE method possesses multi-actor capabilities, enabling it to be applied to various scenarios such as Vehicle-to-Pedestrian (V2P) tests [17]. After successfully passing through MIL, HIL, and VVE testing phases, the autonomous driving algorithm attains the capability to operate in the real world with a higher degree of confidence. This rigorous testing pipeline ensures that the algorithm is robust and ready for the final phase of validation through public road testing.

## Methodology

### *Model-in-Loop and Vehicle Dynamic Model*

The first stage of the proposed testing process is Model-in-the-Loop (MIL) study, where a simulation environment is used to develop a suitable ADAS system for the proposed use case. This would require a vehicle model that can represent the real vehicle behaviors in a reasonably accurate manner. In this section, an extended single-track vehicle model that captures both the longitudinal and lateral dynamic behaviors is introduced. In general, the proposed vehicle model contains three major components: an extended single-track lateral model to capture the longitudinal and lateral dynamics of the vehicle, a Modified Dugoff coupled tire model to provide the tire forces that feed into the single-track model and a wheel rotation model to aid the calculation of tire forces.

The overall model configuration for the extended lateral model is illustrated in Figure 1, which is adapted from [18]. Table 1 lists the parameters used in this model. It can be observed that this model is an extension of the simple single-track lateral model that only makes use of lateral tire forces. Instead, this extended model includes both longitudinal and lateral tire models as well as longitudinal road loads to account for the vehicle dynamic behaviors in both longitudinal and lateral directions.

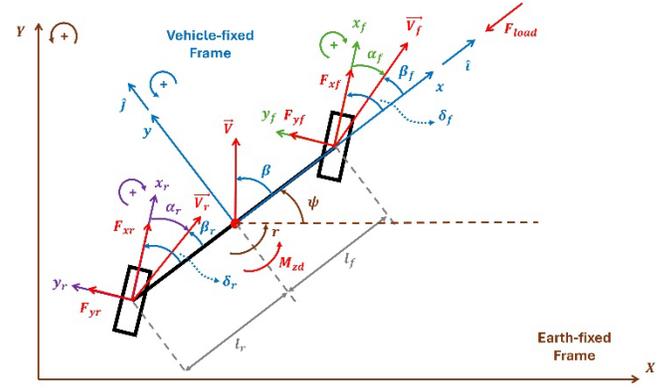

Figure 1. Extended lateral dynamic model

Table 1. Extended lateral dynamic model parameters

| Parameters | Descriptions |
|---|---|
| $m$ | Vehicle mass [kg] |
| $l_f$ | Distance between front axle & CG [m] |
| $l_r$ | Distance between rear axle & CG [m] |
| $I_z$ | Vehicle yaw moment of inertia |
| $V, V_f, V_r$ | Vehicle CG, front & rear axle velocity [m/sec] |
| $\beta, \beta_f, \beta_r$ | Vehicle CG, front & rear axle side slip angle [rad] |
| $\delta_f, \delta_r$ | Front & rear steer angle [rad] |
| $\alpha_f, \alpha_r$ | Front & rear tire side slip angle [rad] |
| $F_{xf}, F_{xr}$ | Front & rear tire longitudinal force [N] |
| $F_{yf}, F_{yr}$ | Front & rear tire lateral force [N] |
| $\psi$ | Vehicle yaw angle [rad] |
| $r$ | Vehicle yaw rate [rad/sec] |
| $M_{zd}$ | Vehicle yaw moment disturbance [Nm] |
| $F_{load}$ | Longitudinal road load [N] |

From Figure 1, it is possible to write down the dynamic equations of vehicle side slip angle ($\beta$), vehicle speed ($V$) and vehicle yaw rate ($r$) as described in Equation 1.

$$\begin{bmatrix} \dot{\beta} \\ \dot{V} \\ \dot{r} \end{bmatrix} = \begin{bmatrix} -\frac{\sin(\beta)}{mV} & \frac{\cos(\beta)}{mV} & 0 \\ \frac{\cos(\beta)}{m} & \frac{\sin(\beta)}{m} & 0 \\ 0 & 0 & \frac{1}{I_z} \end{bmatrix} \begin{bmatrix} \Sigma F_x \\ \Sigma F_y \\ \Sigma M_z \end{bmatrix} - \begin{bmatrix} r \\ 0 \\ 0 \end{bmatrix} \quad (1)$$

Accounting for the resultant forces and moments in the vehicle-fixed frame based on the slightly simplified Figure 2, one can write Equation 2.

$$\begin{bmatrix} \Sigma F_x \\ \Sigma F_y \\ \Sigma M_z \end{bmatrix} = \begin{bmatrix} \cos(\delta_f) & \cos(\delta_r) & -\sin(\delta_f) & -\sin(\delta_r) \\ \sin(\delta_f) & \sin(\delta_r) & \cos(\delta_f) & \cos(\delta_r) \\ l_f \sin(\delta_f) & -l_r \sin(\delta_r) & l_f \cos(\delta_f) & -l_r \cos(\delta_r) \end{bmatrix} \begin{bmatrix} F_{xf} \\ F_{xr} \\ F_{yf} \\ F_{yr} \end{bmatrix} + \begin{bmatrix} -F_{load} \\ 0 \\ M_{zd} \end{bmatrix} \quad (2)$$

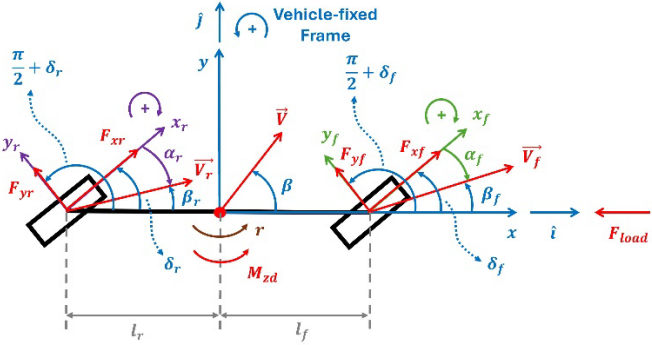

Figure 2. Extended lateral dynamic model represented in vehicle-fixed frame

Combining Equation 1 and Equation 2, one can arrive at the extended model equations of motion (EOM) as illustrated in Equation 3. It should be remarked that this model is essentially a nonlinear longitudinal and lateral single-track dynamic model. Also note that in this model, $(\beta, V, r)$ are regarded as system states, $(\delta_f, \delta_r, M_{zd})$ are treated as system inputs, and tire forces $(F_{xf}, F_{xr}, F_{yf}, F_{yr})$ are to be calculated from the tire model before being fed into this model.

$$\begin{bmatrix} \dot{\beta} \\ \dot{V} \\ \dot{r} \end{bmatrix} = A \begin{bmatrix} F_{xf} \\ F_{xr} \\ F_{yf} \\ F_{yr} \end{bmatrix} + B \begin{bmatrix} -F_{load} \\ 0 \\ M_{zd} \end{bmatrix} - \begin{bmatrix} r \\ 0 \\ 0 \end{bmatrix} \quad (3)$$

where:
$$A = \begin{bmatrix} \frac{\sin(\delta_f - \beta)}{mV} & \frac{\sin(\delta_r - \beta)}{mV} & \frac{\cos(\delta_f - \beta)}{mV} & \frac{\cos(\delta_r - \beta)}{mV} \\ \frac{\cos(\delta_f - \beta)}{m} & \frac{\cos(\delta_r - \beta)}{m} & \frac{-\sin(\delta_f - \beta)}{m} & \frac{-\sin(\delta_r - \beta)}{m} \\ \frac{l_f \sin(\delta_f)}{I_z} & \frac{-l_r \sin(\delta_r)}{I_z} & \frac{l_f \cos(\delta_f)}{I_z} & \frac{-l_r \cos(\delta_r)}{I_z} \end{bmatrix}$$

$$B = \begin{bmatrix} \frac{-\sin(\beta)}{mV} & \frac{\cos(\beta)}{mV} & 0 \\ \frac{\cos(\beta)}{m} & \frac{\sin(\beta)}{m} & 0 \\ 0 & 0 & \frac{1}{I_z} \end{bmatrix}$$

In order to provide the above-mentioned single-track model with tire forces, an appropriate tire model is needed. In this paper, the Modified Dugoff model is used due to its dependency on only a small number of parameters as well as its capability to capture longitudinal and lateral tire force coupling effects. Equation 4 describes the equations for longitudinal and lateral tire forces. Both equations are adapted from [19]. Table 2 details the parameters used in the tire model.

$$\begin{cases} F_x = \frac{C_x s}{1-s} f(Z) g_x \\ F_y = \frac{C_y \tan(\alpha)}{1-s} f(Z) g_y \end{cases} \quad (4)$$

where:
$$\begin{cases} Z = \frac{\mu F_z (1-s)}{2\sqrt{(C_x s)^2 + (C_y \tan(\alpha))^2}} \\ f(Z) = \begin{cases} Z(2-Z) \text{ if } Z < 1 \\ 1 \text{ if } Z \geq 1 \end{cases} \\ g_x = (1.15 - 0.75\mu)s^2 - (1.63 - 0.75\mu)s + 1.5 \\ g_y = (\mu - 1.6)\tan(\alpha) + 1.5 \end{cases}$$

Table 2. Modified Dugoff tire model parameters

| Parameters | Descriptions |
|---|---|
| $C_x$ | Longitudinal tire stiffness [N] |
| $C_y$ | Lateral tire stiffness [N/rad] |
| $s$ | Tire longitudinal slip, $s \in [-1,1]$ |
| $\alpha$ | Tire side slip angle [rad] |
| $F_x$ | Longitudinal tire force |
| $F_y$ | Lateral tire force |
| $\mu$ | Road friction coefficient |
| $F_z$ | Tire vertical load [N] |

It should be remarked that the inputs to the tire model are tire longitudinal slip ($s$) and lateral side slip angle ($\alpha$). While the tire side slip angle can be obtained by applying post-processing on the outputs of the extended lateral model mentioned above, the calculation of tire longitudinal slip requires an additional model describing the wheel rotational dynamics. This wheel rotational model is illustrated in Figure 3 and its dynamic equations are listed in Equation 5 and Equation 6. Table 3 lists the necessary parameters for this model. Note that the deviated angular velocities of the front and rear tire ($\Delta \omega_f, \Delta \omega_r$) are introduced to achieve wheel-vehicle speed synchronization and to avoid small amplitude oscillations in the longitudinal tire forces for undriven wheels, which can cause numeric issues during simulation, especially at low speed.

$$\begin{cases} I_f \Delta \dot{\omega}_f = M_f - F_{xf} R_f \\ I_r \Delta \dot{\omega}_r = M_r - F_{xr} R_r \end{cases} \quad (5)$$

where: $\Delta \omega_{f0} = \Delta \omega_{r0} = 0$

$$\begin{cases} \omega_f = \Delta \omega_f + \frac{V_{fxf}}{R_f} \\ \omega_r = \Delta \omega_r + \frac{V_{rxr}}{R_r} \end{cases} \quad (6)$$

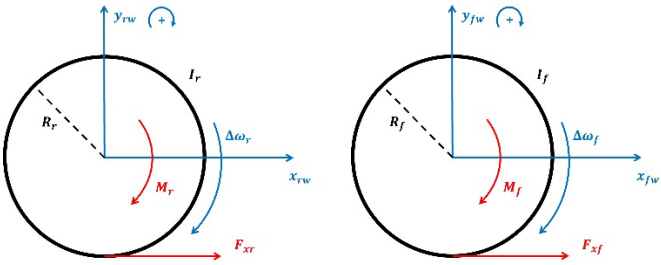

Figure 3. Wheel rotation model

Table 3. Wheel rotation model parameters

| Parameters | Descriptions |
|---|---|
| $\omega_f, \omega_r$ | Front & rear tire angular velocity [rad/sec] |
| $\Delta\omega_f, \Delta\omega_r$ | Front & rear tire deviated angular velocity [rad/sec] |
| $R_f, R_r$ | Front & rear tire radius [m] |
| $I_f, I_r$ | Front & rear tire moment of inertia |
| $M_f, M_r$ | Front & rear tire driving torque [Nm] |
| $V_{fxf}$ | Front axle longitudinal velocity in front wheel frame [m/sec] |
| $V_{rxr}$ | Rear axle longitudinal velocity in rear wheel frame [m/sec] |

Combining all the above-mentioned components, one can construct the overall vehicle model as illustrated in Figure 4. It can be observed that vehicle position $(X, Y)$ and orientation/yaw angle $(\psi)$ can be obtained through model output post-processing.

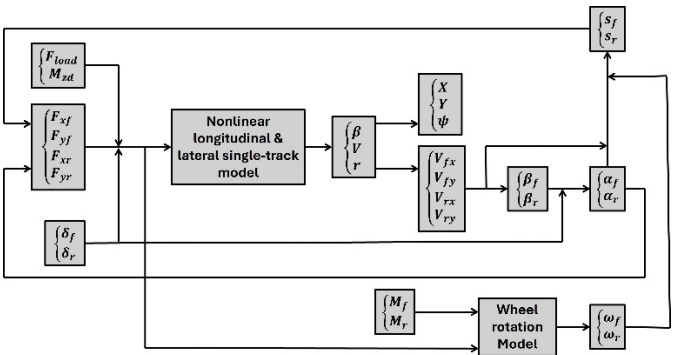

Figure 4. Overall vehicle model structure

## Hardware-in-Loop Test

Hardware-in-the-Loop (HIL) testing is an exceptionally effective simulation and validation technique widely employed in the development and testing of autonomous driving functions. The core principle of HIL testing involves integrating actual hardware components in vehicle, such as autonomous driving controllers (MicroAutobox), GPS, and dedicated short-range communications (DSRC), etc., into the simulation environment. This is very close to real vehicle testing except the real vehicle is replaced by a vehicle model that can accurately simulate vehicle dynamics. This setup allows for the evaluation of hardware performance of autonomous driving functions under pre-set operating conditions in a laboratory. Additionally, HIL testing enables the evaluation of autonomous driving functions in real-time scenarios and under signal transmission delays, while also facilitating the detection of unrealistic extreme inputs that could affect system reliability. By conducting all experiments in the laboratory, HIL testing significantly reduces overall testing costs and avoids the risks associated with on-road trials. Furthermore, HIL testing serves as an indispensable precursor to public road testing, ensuring that autonomous driving systems are thoroughly evaluated for safety and performance before being deployed in real-world environments.

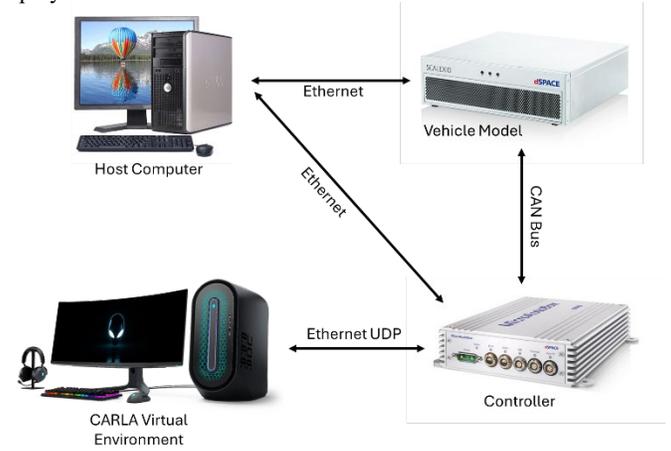

Figure 5. Hardware-in-the-Loop (HIL) test setup

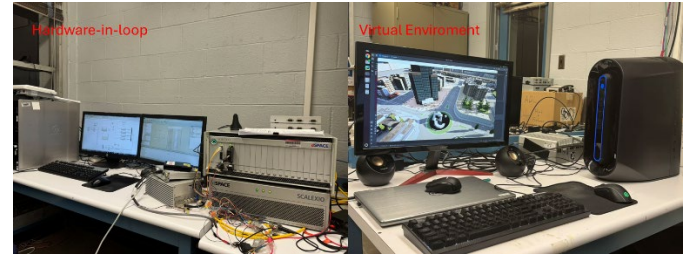

Figure 6. Hardware layout in the lab

Figure 5 demonstrates the Hardware-in-the-Loop (HIL) test setup employed in this paper. Figure 6 illustrates our laboratory's hardware configuration, which includes a HIL simulation workstation, and a computer dedicated to managing the virtual environment. In our configuration, the aforementioned vehicle dynamic Simulink model operates on a Scalexio simulation computer in real time, continuously calculating the vehicle's position and state based on inputs such as throttle, brake, and steering. These vehicle states are transmitted via the CAN bus to the MicroAutobox (MABX) electronic control unit, which functions both as the vehicle's controller and as a communication hub between the vehicle model and the virtual environment.

Then, the MABX sends the current vehicle information back to the computer managing the virtual environment and receives environmental data in return. It processes this data to generate appropriate vehicle control commands. Communication between the MABX and the virtual environment computer is developed through Ethernet UDP. The virtual environment computer then uses the data from the MABX to update the vehicle's status within the simulated environment, providing a visual representation of the experiment. Additionally, the Host Computer controls the operations of both the Scalexio and MABX systems, coordinating the overall simulation process.

This HIL setup establishes an integrated, real-time feedback loop where the vehicle dynamic model and corresponding virtual environment interact dynamically. This enables comprehensive testing of the vehicle control system under various simulated conditions without the need for physical road tests. As a result, HIL testing not only enhances safety by minimizing risks associated with on-road trials but also significantly reduces the time and costs involved in prototype development and verification.

Moreover, the HIL test setup is designed for seamless transition to real vehicle including VVE testing. In real-world scenarios, the Scalexio simulation computer can be replaced with an actual vehicle, and the computer managing the virtual environment can be substituted with an in-vehicle PC capable of reading sensor data and analyzing the traffic environment. Additionally, the existing HIL architecture can be enhanced by integrating various other hardware components to increase the realism and reliability of the simulation. For example, DSRC or other communication modules can be added to simulate Vehicle-to-Vehicle (V2V) or Vehicle-to-Pedestrian (V2P) communications, roadside units (RSU) can be incorporated to transmit real infrastructure data to the HIL system, GPS modules can be included to obtain accurate GPS signals and traffic control computers can be added to generate signal phase and timing (SPaT) and MAP messages. These enhancements allow for a more comprehensive and realistic testing environment, further bridging the gap between simulated and real-world autonomous driving scenarios. By incorporating these additional hardware elements, the HIL setup becomes a versatile and robust platform for both development and extensive validation of autonomous driving systems before their deployment on public roads.

## *Vehicle-In-Virtual-Environment Test*

Given that Hardware-in-the-Loop (HIL) employs all the real equipment necessary for ADAS testing except for the real vehicle, the natural next step would be to commence testing that involves the real vehicle, currently the most common approach of which is to perform public road testing. Carrying out the test on public road, however, comes with several critical drawbacks. Firstly, other road users are involuntarily involved in the testing of the ADAS systems, which poses safety concerns, especially during extreme and edge cases. Secondly, the previously mentioned extreme and edge cases are typically rare occurrences, and would require significant milage to encounter and test, limiting the overall efficiency of the approach. In that regard, the Vehicle-in-Virtual-Environment (VVE) was proposed as a novel approach to perform real-vehicle testing in a safe and efficient manner.

The overall architecture of the VVE approach is displayed in Figure 7. The real vehicle is operated in a safe and open testing space, where its motions are synchronized, via frame transformation, with those of a virtual vehicle operating in a highly realistic virtual environment. Depending on the desired traffic scenario to be tested, the virtual environment can be edited with relative ease. Virtual onboard sensors of our choice can be fitted to the virtual vehicle and its data collected from the virtual environment can be fed into the onboard equipment of the real vehicle so that the real vehicle control unit can react to a virtual traffic scenario.

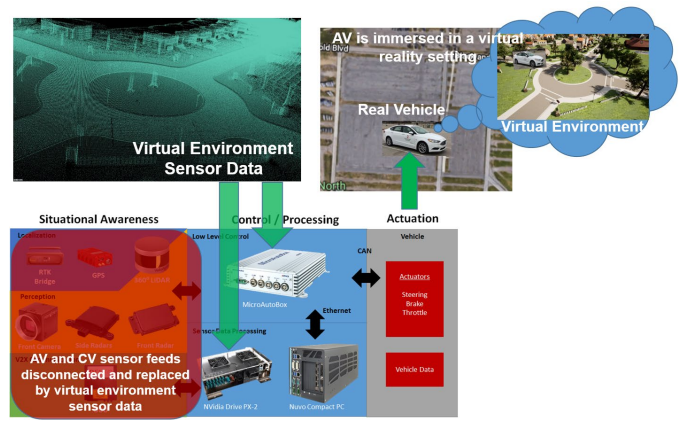

Figure 7. VVE Architecture [16]

With the synchronization in place, difficult, rare and safety critical tests can be conducted by simply editing the virtual environment to create the respective traffic scenarios, reducing the cost of testing. This method is also significantly safer, as the vehicle operates in a separate open space and does not run the risk of a real traffic accident should the ADAS system being tested fail the safety critical experiment. The utilization of real vehicle dynamics is another benefit of the VVE approach, as this simulation-like approach combines the benefits of real vehicle testing without invoking any safety and cost drawbacks. This approach also provides the possibility of multi-actor experiments. Figure 8 provides an example for this type of test in the form of a Vehicle-to-Pedestrian (V2P) communication-based collision avoidance experiment. The real pedestrian and the real vehicle operate in separate spaces that are safe and open. The real pedestrian is equipped with a mobile phone that has IMU and GPS sensors installed as well as a mobile application that calculates the heading and position of the pedestrian, and this motion information is broadcasted via Bluetooth low-energy (BLE) connection. This pedestrian motion data, together with the vehicle motion data, are synchronized through frame transformation to a virtual pedestrian and a virtual vehicle operating within the same virtual environment where virtual vehicle-to-pedestrian collision is possible. V2P-based collision avoidance experiment can hence be carried out under this setup, where the virtual vehicle feeds the virtual pedestrian motion data into the real vehicle so that the real vehicle can react to avoid collisions in the virtual world.

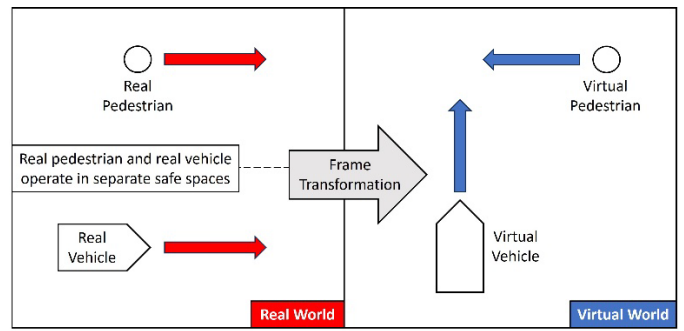

Figure 8. V2P test using VVE approach [16]

Figure 9 illustrates the current implementation structure of the corresponding VVE setup, and Figure 10 displays the equipment onboard our test vehicle. The real vehicle has a RTK GPS unit with differential antennas that provides us with both vehicle position and heading information, and this data is fed into the dSpace microautobox (MABX) unit, which is our onboard control unit, before being sent via Ethernet UDP protocol to an in-vehicle PC that runs an Unreal Engine-based CARLA virtual environment. The virtual environment applies the frame transformation routine to the received data to achieve the desired real-virtual synchronization. On the other hand, the virtual sensor data collected in the virtual environment is fed back into the MABX unit using Ethernet UDP protocol again.

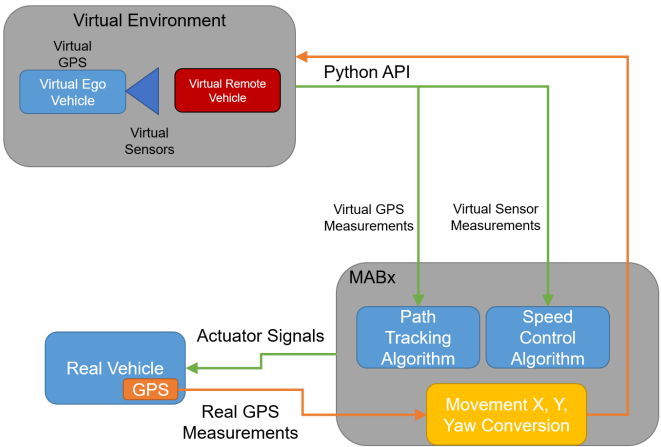

Figure 9. Implementation structure of VVE approach [16]

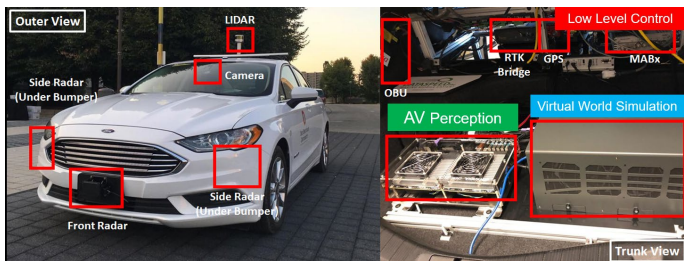

Figure 10. Test vehicle used for VVE approach [16]

# Experiments and Results

## *Model-in-Loop Test*

In this section, we present how to use MIL to train and evaluate a Deep Reinforcement Learning (DRL)-based collision avoidance agent. By employing a vehicle dynamic model within the MIL setup (as demonstrated earlier), which closely mirrors the dynamics of a real vehicle, the training results are highly reliable and can be easily applied to HIL and VVE testing environments. This close resemblance ensures that the training outcomes are very similar to real-world scenarios and are thus highly dependable. The first part of the section provides a detailed explanation of how to transform autonomous driving task in a complex traffic environment into a Markov Decision Process (MDP) problem. We delve into the specific structure and algorithms of the DRL approach and demonstrate how the DRL-based collision avoidance agent addresses this problem. The second part focuses on effectively integrating the DRL algorithm and training process with MIL. We discuss how to seamlessly combine the vehicle model in Simulink with the CARLA virtual environment to train the agent successfully. These training results will also be further utilized in HIL and VVE testing in future work.

## **Deep Reinforcement Learning Based Collision Avoidance**

Path planning and collision avoidance algorithms have already become important components of Automated Driving Systems (ADS) and significantly affect the overall performance of autonomous vehicles. Unlike traditional automotive navigation systems, autonomous driving path planning must consider not only the overall route from the starting point to the destination but also local collision avoidance during path tracking. Therefore, the purpose of this section is to design a reliable and robust path planning and collision avoidance algorithm for ADS, which can help enhance the overall performance of autonomous vehicles and elevate their SAE autonomous driving level.

Currently, in the field of autonomous driving path planning and decision-making, there exist numerous research that use either traditional modular and optimization methods or machine learning based approaches. Traditional methods often employ a modular framework, breaking down the autonomous driving problem into several modules including perception, planning, decision-making, and control. This modularity allows for easier implementation and adaptability. Optimization-based techniques like Elastic Band [20], Potential Field [21], and SVM-based [22] optimization are well-understood and can perform effectively in simple environments. Advantages of these traditional methods include their ease of implementation, good performance in simple traffic environments, low computational requirements enabling real-time performance, and being intuitive which make its principle relatively easy to explain. However, they also have significant disadvantages, such as control infeasibility, a tendency to get stuck in local minima, and computational inefficiency in complex environments. Additionally, methods like Control Lyapunov Functions (CLF) and Control Barrier Functions (CBF) [23-25] face challenges in designing effective constraints, which limits their applicability to oversimplified vehicle dynamic models.

Given these deficiencies of the traditional approach, machine learning approach, particularly Deep Reinforcement Learning (DRL) based method, emerges as a promising alternative. Unlike traditional optimization methods, DRL based approach treats the path planning and collision avoidance problem as a Markov Decision Process (MDP), allowing it to optimize the entire decision-making process. DRL based autonomous driving agent can learn from interaction in complex traffic environments, adapt to various traffic scenarios, and satisfy real-time performance requirements, making it a more robust and advanced solution. While training a high-performance DRL agents requires large computational resources and is highly dependent on training data quality, its ability to continuously improve through interaction with the environment makes it ideal for addressing the limitations of conventional approaches. Kendall et al. were among the first to apply DRL to autonomous driving, proposing an end-to-end model that demonstrated the potential of this approach [26]. Following this pioneering work, numerous researchers have explored various DRL architectures and Markov Decision Process (MDP) designs for autonomous vehicles. Yurtsever et al. developed a hybrid DRL framework for Automated Driving Systems (ADS) [27]. Aksjonov et al. combined traditional rule-based approaches with machine learning to enhance autonomous driving capabilities [28]. Makantasis et al. introduced a DDQN-based driving policy adaptable to mixed traffic environments, testing its efficacy across different market penetration levels [29]. Nageshrao et al. integrated DDQN with a short-horizon safety mechanism for highway scenarios under varying traffic densities [30]. Peng et al. utilized a Dueling Double Deep Q-Network (DDDQN) framework in an end-to-end ADS, validating its efficiency using The Open Racing Car Simulator (TORCS) [31]. Jaritz et al. proposed an Asynchronous Actor-Critic (A3C) method that maps front camera images to driving actions, demonstrating faster convergence and robust performance [32]. To handle critical pre-accident situations, Merola et al. applied a DQN-based approach for executing emergency maneuvers to minimize or avoid damage [33]. Cao et al. developed a hierarchical reinforcement and imitation learning (H-REIL) approach to balance safety and efficiency in near-accident scenarios [34].

Building upon these advancements, we introduce a novel Double Deep Q-Network (DDQN) based autonomous driving agent designed to enhance collision avoidance performance by learning to brake in emergency traffic conditions. DDQN is an improvement over the Deep Q-Network (DQN), which was also a significant breakthrough in the reinforcement learning field. DQN, introduced by Mnih et al from DeepMind Technologies in 2013 [30-31], successfully combined Q-learning with deep learning to handle state spaces with multiple dimensions, incorporating key innovations such as experience replay and target networks. However, DQN suffered from overestimation of action values due to the maximization bias in the Q-learning update, leading to suboptimal policies and unstable learning. To address this issue, DDQN was developed by Hasselt et al. in 2015 while working at DeepMind. DDQN separates the action selection process from target Q-value estimation, mitigating overestimation bias and enhancing learning stability [37]. As a result, DDQN is expected to outperform traditional methods in collision avoidance scenarios.

For an autonomous vehicle, performing collision avoidance under emergency traffic condition is a dynamic, ongoing decision-making task that is very similar to a Markov Decision Process (MDP). The autonomous vehicle continuously monitors its surroundings and its own operational status, enabling it to make quick and precise decisions to navigate safely based on its own location and speed. To effectively manage the uncertainties, present in traffic environments, DDQN is proposed to develop an autonomous driving agent that optimizes decisions for maximum safety and efficiency. Within the MDP framework, the state space contains an occupancy grid that maps surrounding obstacles, ego vehicle's status, waypoints of pre-calculated path, and detailed obstacle information such as time-to-collision-zones (TTZ). The action space includes a range of discrete control commands, including steering, throttle, and braking, tailored to respond adaptively to various driving conditions. The transition model contains the aforementioned vehicle dynamic model and the CARLA virtual traffic environment to predict subsequent states based on current actions, while the reward function evaluates immediate feedback reward of state transitions to guide the autonomous vehicle towards safer and more efficient driving behaviors.

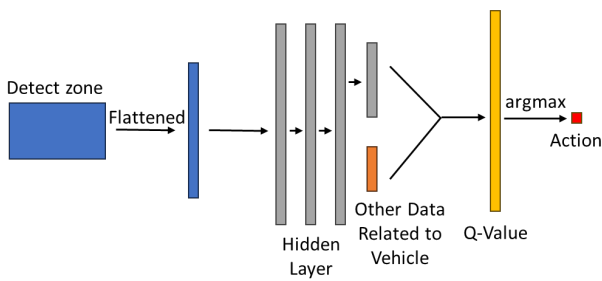

Figure 11. Proposed DDQN neural network structure [38]

```
Algorithm 1
1:  Initialize replay memory D
2:  Initialize target network Q̂ and Online Network Q with random weights θ
3:  for each episode do
4:      Initialize traffic environment
5:      for t = 1 to T do
6:          With probability ε select a random action a_t
7:          Otherwise select a_t = max_a Q*(s_t, a; θ)
8:          Execute a_t in CARLA and extract reward r_t and next state s_{t+1}
9:          Store transition (s_t, a_t, r_t, s_{t+1}) in D
10:         if t mod training frequency == 0 then
11:             Sample random minibatch of transitions (s_j, a_j, r_j, s_{j+1}) from D
12:             Set y_j = r_j + γ max_{a_{j+1}} Q̂(s_{j+1}, argmax_{a_{j+1}} Q(s_j, a_{j+1}; θ); θ)
13:             for non-terminal s_{j+1}
14:             or y_j = r_j for terminal s_{j+1}
15:             Perform a gradient descent step to update θ
16:             Every N steps reset Q̂ = Q
17:         end if
18:         Set s_{t+1} = s_t
19:     end for
20: end for
```

Figure 12. Proposed DDQN autonomous driving algorithm [38]

Figures 11 and 12 illustrate the neural network architecture of the proposed DDQN-based autonomous driving agent. The network consists of four layers: the pre-processed input data first pass through three hidden layers, each containing 128 neurons, and then proceed to a final layer with 32 neurons. In the pre-processing stage, we begin by flattening the occupancy grid to convert spatial data into a one-dimensional vector. This flattened vector is then sequentially processed through the three hidden layers, which extract high-level features from the input data.

After these layers, we integrate additional sensor information— including the ego vehicle's status, path tracking waypoints, and data about other road users—with the output from the third hidden layer. This combined data is fed into the final 32-neuron layer, which computes the Q-values for all possible actions. This architectural design is intentional; it preserves critical information that might be lost in earlier layers, ensuring that the network effectively considers important details in its decision-making process.

The DDQN framework contains two neural networks, online network $Q$ for actions generation and target network $\hat{Q}$ for target $Q$-values computation, during training. The parameter of target neural network only updates at specified intervals. This decoupling mitigates overestimation bias and stabilizes the training process by preventing rapid changing in the learning targets.

The aforementioned vehicle dynamic model operates within the CARLA traffic simulation environment to generate training data for the DDQN agent, which are stored in a replay buffer. This buffer breaks the temporal correlations between consecutive training samples and reduces variance during the training process. By randomly sampling from the replay buffer, we enhance data utilization efficiency and ensure balanced and comprehensive learning throughout the training phases. This approach allows the agent to generalize better across a variety of traffic scenarios, ultimately improving collision avoidance performance.

## DRL with MIL Training Setting

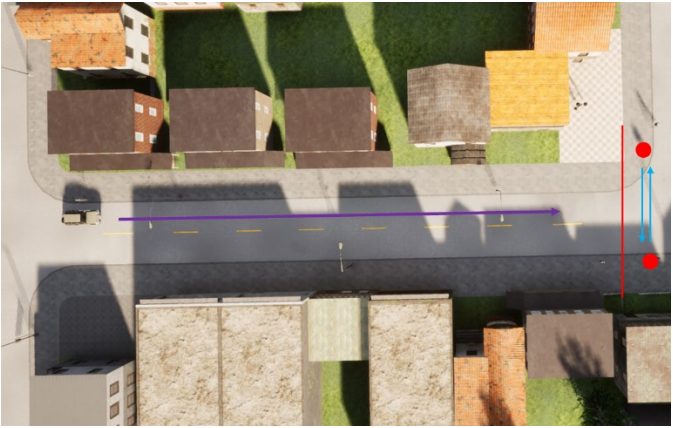

Figure 13. Traffic scenario used for testing.

Figure 13 demonstrates the traffic scenario within the virtual environment used to evaluate the overall performance of the DDQN-based autonomous driving agent. In this scenario, the vehicle navigates along a street while two pedestrians cross the crosswalk and walk back and forth. The vehicle is tasked with maintaining a comfortable speed and coming to a safe stop before the crosswalk when pedestrians are present. This test scenario is specifically designed to evaluate the agent's ability to perform emergency braking maneuvers effectively when encountering unexpected pedestrian movements. By simulating such dynamic and potentially hazardous conditions, this scenario ensures that the DDQN agent can reliably respond to emergencies, demonstrating its capability to maintain safety and control in various traffic situations.

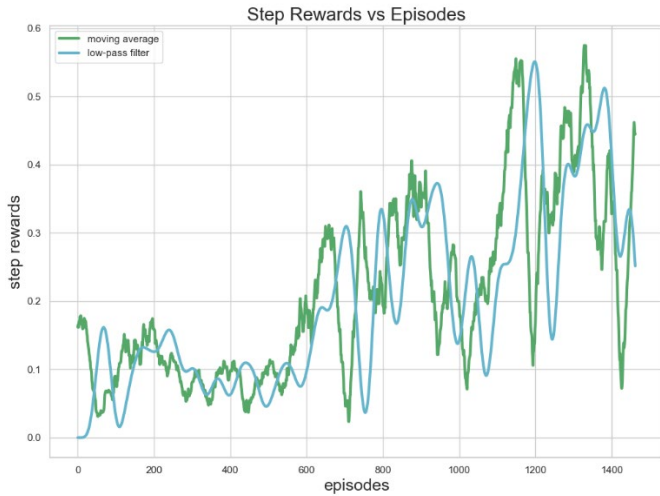

Figure 14. DDQN autonomous driving agent reward progress

Figure 14 illustrates the training reward progression of the DDQN-based autonomous driving agent over 1,500 episodes. At the initial stage of training, the average reward per step is notably low, which is expected as the agent initially selects actions randomly from the action space to explore all possibilities within the virtual environment. As training continues, the average step reward steadily increases, reflecting the agent's learning and improvement in decision-making. By approximately episode 800, the average reward stabilizes around 0.3, indicating that the agent has converged to an optimized policy. This stabilization demonstrates that the DDQN has effectively learned to select actions that consistently maximize cumulative rewards, validating the success of the training process in enhancing the agent's brake capabilities in emergency traffic condition. Overall, the reward progression plot confirms that the DDQN training effectively transitions the agent from exploration to exploitation, resulting in a robust and reliable autonomous driving policy.

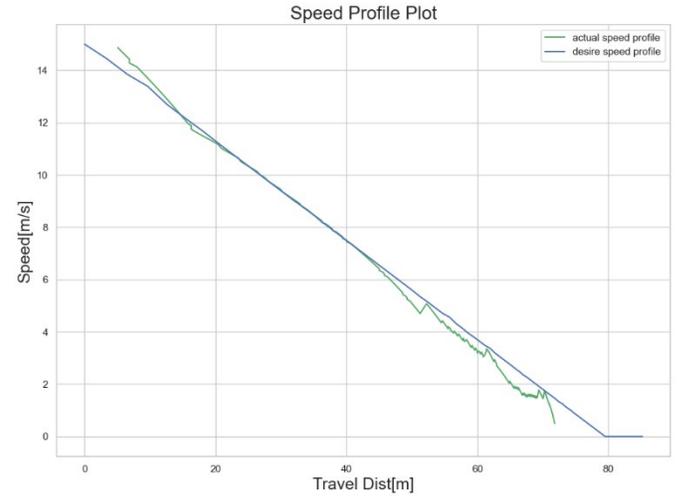

Figure 15. DDQN autonomous driving agent speed tracking plot with initial speed 15m/s

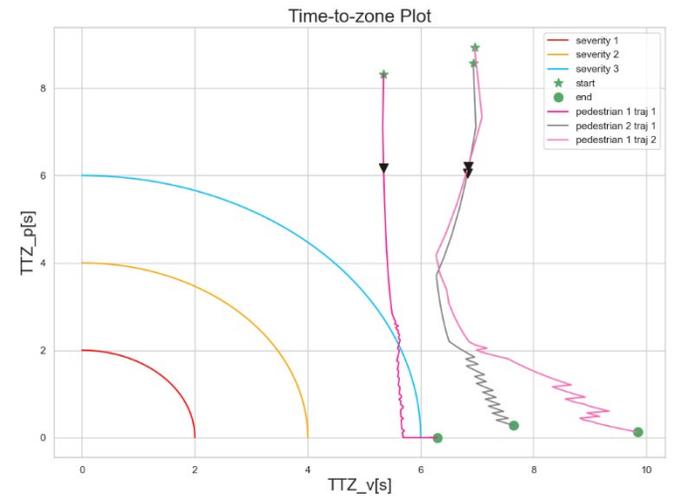

Figure 16. DDQN autonomous driving agent TTZ plots with initial speed 15m/s

Figure 15 presents the speed tracking performance of the DDQN-based autonomous driving agent, starting with an initial speed of 15 m/s. Initially, the vehicle maintains the set speed of 15 m/s and gradually deaccelerates to follow the desired braking profile which is decreasing its speed smoothly from 15 m/s to 0 m/s. However, at lower speeds, the vehicle exhibits some abrupt braking behavior, resulting in less precise tracking of the desired braking speed. This occurs because the vehicle's position is in very close to the pedestrians, prompting the agent to brake more aggressively to avoid potential collisions. Such behavior demonstrates the agent's prioritization of safety by ensuring timely and sufficient deceleration when immediate threats are detected.

Figure 16 illustrates the Time-To-Collision (TTZ) performance for the DDQN autonomous driving agent, starting with an initial speed of 15 m/s. In the TTZ plots, red circles indicate both pedestrian and vehicle's TTZ times less than 2 seconds (indicating very urgent situations), orange circles represent both pedestrian and vehicle's TTZ times less than 4 seconds, and blue circles represent both pedestrian and vehicle's TTZ times less than 6 seconds. Throughout the entire duration of the simulation, the TTZ correspondence between the vehicle and the two pedestrians consistently remains above 4 seconds, and often exceeds 6 seconds. This consistent maintenance of safe TTZ values highlights the effectiveness of the DDQN agent in ensuring pedestrian safety, even in emergency scenarios. By keeping socially acceptable distance [39] and reacting promptly to dynamic obstacles, the DDQN agent successfully minimizes the risk of collisions, demonstrating its capability to handle critical situations reliably.

### *Hardware-in-Loop Test*

Figure 17 demonstrates the speed tracking performance of the DDQN-based autonomous driving agent during HIL simulation tests, starting with an initial speed of 15 m/s. Overall, the trained model successfully tracks the desired speed profile and comes to a complete stop before the crosswalk under real-time HIL simulation conditions. Consistent with the MIL tests, the agent exhibits a sudden increase in braking intensity at lower speeds, enabling the vehicle to decelerate rapidly and halt to ensure pedestrian safety.

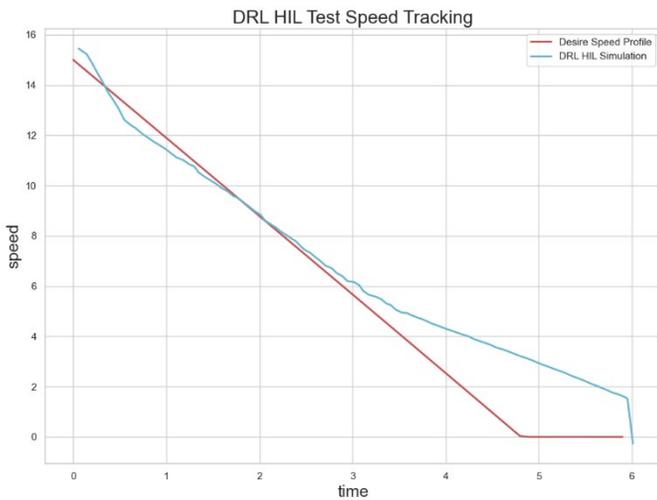

Figure 17. DDQN autonomous driving agent HIL simulation speed tracking plot with initial speed 15m/s

During the HIL tests, we evaluated the model's performance with the integration of actual hardware, and the results were highly satisfactory. The agent maintained accurate speed tracking and demonstrated reliable emergency braking behavior, even when interacting with physical components of the system. This positive performance in the HIL environment indirectly confirms the viability of our proposed development and testing pipeline. The successful integration and real-time responsiveness of the DDQN agent in the HIL setup indicates that our approach is both feasible and effective for advancing autonomous driving systems.

### *Vehicle-In-Virtual-Environment Test*

In VVE testing, we evaluated the synchronization between a real vehicle operating in a physical environment and its virtual counterpart by conducting manual driving tests. Figures 18(a) and 18(b) display the vehicle trajectories in the virtual and real environments, respectively. It is evident that the trajectories perfectly overlap, demonstrating that our vehicle dynamics are accurately replicated and transmitted to the virtual environment. This precise synchronization ensures the reliability and accuracy of VVE testing, validating that the virtual simulations accurately represent real-world vehicle behavior. Due to time constraints, additional tests will be conducted and progressively incorporated into the paper to further prove and evaluate the VVE framework. This initial success provides a strong foundation for the continued development and refinement of our VVE testing methodology, ensuring robust and comprehensive evaluation of autonomous driving systems in the future.

## Conclusion

In order to address challenges caused by urbanization and in order to address challenges caused by urbanization and the increasing number of privately owned vehicles which is the leading reason of traffic congestion and a surge in car accidents, it is important to find effective solutions to enhance road safety. Automated Driving Systems (ADS) have shown significant promise in reducing accidents caused by human errors, which account for approximately 75% of all car accidents globally. Traditional testing pipelines, such as Model-in-the-Loop (MIL) and Hardware-in-the-Loop (HIL) simulations followed by public road deployment, have proven inadequate due to safety risks, high costs, and inefficiencies. To overcome these limitations, this paper has proposed a novel and comprehensive testing pipeline integrating MIL, HIL, Vehicle-in-Virtual-Environment (VVE) testing, before public road deployment. This approach not only ensures thorough validation of autonomous driving algorithms but also enhances safety, reduces costs, and accelerates development, paving the way for more reliable and effective ADS deployment.

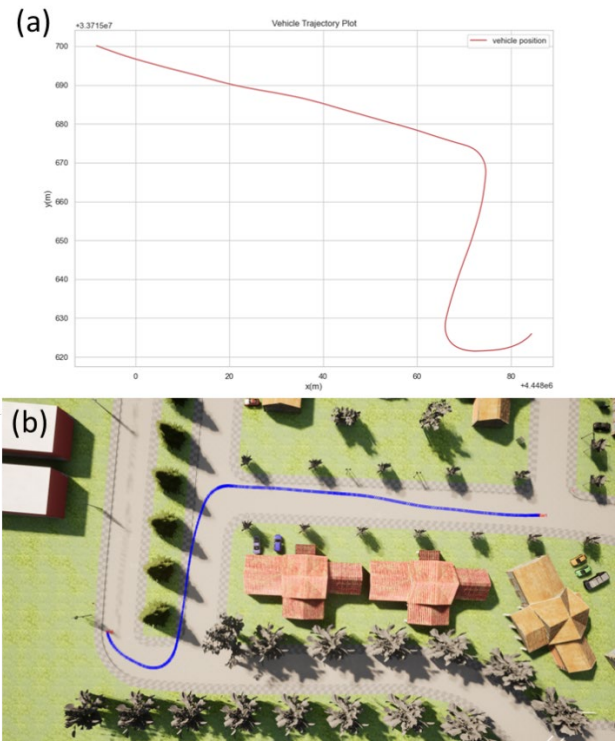

Figure 18. VVE motion synchronization test results: (a) Motion trajectory in the real world; (b) Motion trajectory in the virtual world [16]

In this paper, we demonstrated the principles and implementation of Model-in-the-Loop (MIL), Hardware-in-the-Loop (HIL), and Vehicle-in-Virtual-Environment (VVE) methods for testing autonomous driving algorithms. Using the MIL approach, we trained and evaluated a Double Deep Q-Network (DDQN)-based collision avoidance autonomous driving agent. We then showcased how to evaluate traditional path-following control algorithms using the HIL setup. Furthermore, we demonstrated the VVE method's ability to synchronize real-world and virtual environments, enabling vehicle-to-pedestrian synchronization and collision avoidance testing.

All these tests confirm the feasibility and effectiveness of our proposed experimental methods. By employing this comprehensive testing pipeline, we can efficiently develop advanced and robust autonomous driving systems using artificial intelligence and other cutting-edge algorithms. This approach allows us to thoroughly test the effectiveness of autonomous driving systems under various traffic conditions. Due to time constraints, a full test of the deep reinforcement learning (DRL)-based collision avoidance algorithm is still in progress and will be included in the updated version of the paper.

## Future Work

Despite the advancements presented in this paper, the proposed testing pipeline still has several limitations that need to be addressed in future work. One significant issue is the presence of singular points in the vehicle dynamic model used during Model-in-the-Loop (MIL) and Hardware-in-the-Loop (HIL) testing. Specifically, when the vehicle speed is exactly zero or when input values are extremely large, the simulated data may become inaccurate. This limitation restricts the effectiveness of MIL testing. To overcome this challenge, future efforts will focus on refining the vehicle dynamics model to eliminate these singularities and improve simulation accuracy.

We chose to use the bicycle model in this paper because our primary objective was to demonstrate the effectiveness of the proposed testing pipeline rather than to achieve high-fidelity vehicle dynamics simulation. However, we acknowledge that for more accurate and realistic simulations, a more advanced vehicle dynamics model would be beneficial. In the future, we plan to explore and incorporate more sophisticated vehicle dynamics models to enhance simulation accuracy. Additionally, there are several mature vehicle dynamics simulation tools, such as CarSim [40], that provide detailed and reliable vehicle dynamic simulations. We intend to leverage these technologies in future work to further refine our autonomous driving system development and testing process, ensuring both high accuracy and robustness.

Furthermore, HIL testing currently faces coordination problems; the virtual environment and the vehicle model cannot initiate simultaneously, resulting in minor discrepancies in test results. To address this, we plan to introduce an interactive switch via communication protocols to synchronize the startup of the virtual environment and the vehicle model, thereby enhancing the consistency and reliability of HIL testing.

In addition to these improvements, future research will conduct a thorough analysis of the effects of communication latency and computation delays on the Vehicle-in-Virtual-Environment (VVE) method. Understanding these impacts is crucial for optimizing system performance and ensuring real-time responsiveness. Moreover, future work will explore the inclusion of other types of road users, such as vehicular traffic, to expand the range of testing scenarios. Furthermore, integrating extended reality (XR) goggles into the testing system is another aspect we intend to explore and our preliminary analysis and exploration in this area was very promising. By immersing real pedestrians or drivers into the virtual environment, XR technology can further enhance the realism of testing scenarios, providing more authentic interactions and valuable data for refining autonomous driving algorithms. Vehicle-to-Everything (V2X) communications is being used more widely in automotive and connected and autonomous driving [41], [42] and our future work will also include the use of V2X communication to enhance the situational awareness of our ego vehicle. Future work can also focus on combining learning based approached with well established methods like the elastic band for collision avoidance [43-46].

## Acknowledgments

This project is funded in part by Carnegie Mellon University's Safety21 National University Transportation Center, which is sponsored by the U.S. Department of Transportation.

The authors thank NVIDIA for its GPU donations.

The authors thank the Automated Driving Lab at the Ohio State University.